\documentclass[10pt,twocolumn,letterpaper]{article}

\usepackage{cvpr}              % To produce the CAMERA-READY version
\usepackage{times}
\usepackage{wrapfig}
\usepackage{epsfig}
\usepackage{graphicx}
\usepackage{amsmath}
\usepackage{amssymb}
\usepackage{booktabs}
\usepackage{makecell}
\usepackage{array}
\usepackage{enumitem}

\usepackage{float}
\usepackage{dblfloatfix}    % To enable figures at the bottom of page

\usepackage[accsupp]{axessibility}  % Improves PDF readability for those with disabilities.

\usepackage[pagebackref=true,breaklinks=true,letterpaper=true,colorlinks,bookmarks=false]{hyperref}

\long\def\ignorethis#1{}

\usepackage{color}
\definecolor{gray}{rgb}{0.35,0.35,0.35}
\definecolor{red}{rgb}{1,0,0}
\definecolor{dark-green}{rgb}{0,0.4,0}
\definecolor{blue}{rgb}{0,0,1}
\definecolor{orange}{rgb}{1,0.55,0}
\definecolor{white}{rgb}{1,1,1}
\definecolor{black}{rgb}{0,0,0}
\definecolor{dark-brown}{rgb}{0.2,0.1,0}
\usepackage{xcolor}

\newbox\jsavebox

\usepackage[capitalize]{cleveref}
\crefname{section}{Sec.}{Secs.}
\Crefname{section}{Section}{Sections}
\Crefname{table}{Table}{Tables}
\crefname{table}{Tab.}{Tabs.}

\def\cvprPaperID{35} % *** Enter the CVPR Paper ID here
\def\confName{CVPR}
\def\confYear{2022}

\begin{document}

%%%%%%%%% TITLE 
% \title{Transformaly - Exploit Vision Transformer Power for Anomaly Detection}
\title{Transformaly - Two (Feature Spaces) Are Better Than One}

\author{Matan Jacob Cohen\\
Blavatnik School of Computer Science\\
Tel-Aviv University\\
Tel-Aviv, Israel\\
{\tt\small matanyaakovc@mail.tau.ac.il}

\and
Shai Avidan\\
School of Electrical Engineering\\
Tel-Aviv University\\
Tel-Aviv, Israel\\
{\tt\small avidan@tauex.tau.ac.il}
}
\maketitle

\begin{abstract}
    Anomaly detection is a well-established research area that seeks to identify samples outside of a predetermined distribution. An anomaly detection pipeline is comprised of two main stages: (1) feature extraction and (2) normality score assignment. Recent papers used pre-trained networks for feature extraction achieving state-of-the-art results. However, the use of pre-trained networks does not fully-utilize the normal samples that are available at train time. This paper suggests taking advantage of this information by using teacher-student training. In our setting, a pre-trained teacher network is used to train a student network on the normal training samples. Since the student network is trained only on normal samples, it is expected to deviate from the teacher network in abnormal cases. This difference can serve as a complementary representation to the pre-trained feature vector. Our method - $Transformaly$ - exploits a pre-trained Vision Transformer (ViT) to extract both feature vectors: the pre-trained (agnostic) features and the teacher-student (fine-tuned) features. We report state-of-the-art AUROC results in both the common unimodal setting, where one class is considered normal and the rest are considered abnormal, and the multimodal setting, where all classes but one are considered normal, and just one class is considered abnormal\footnote{The code is available at \url{https://github.com/MatanCohen1/Transformaly}}.
    %  \footnote{Code will be published upon acceptance.}
    
\end{abstract}

\section{Introduction}

\begin{figure}[t]
\begin{center}
\includegraphics[width=1\linewidth]{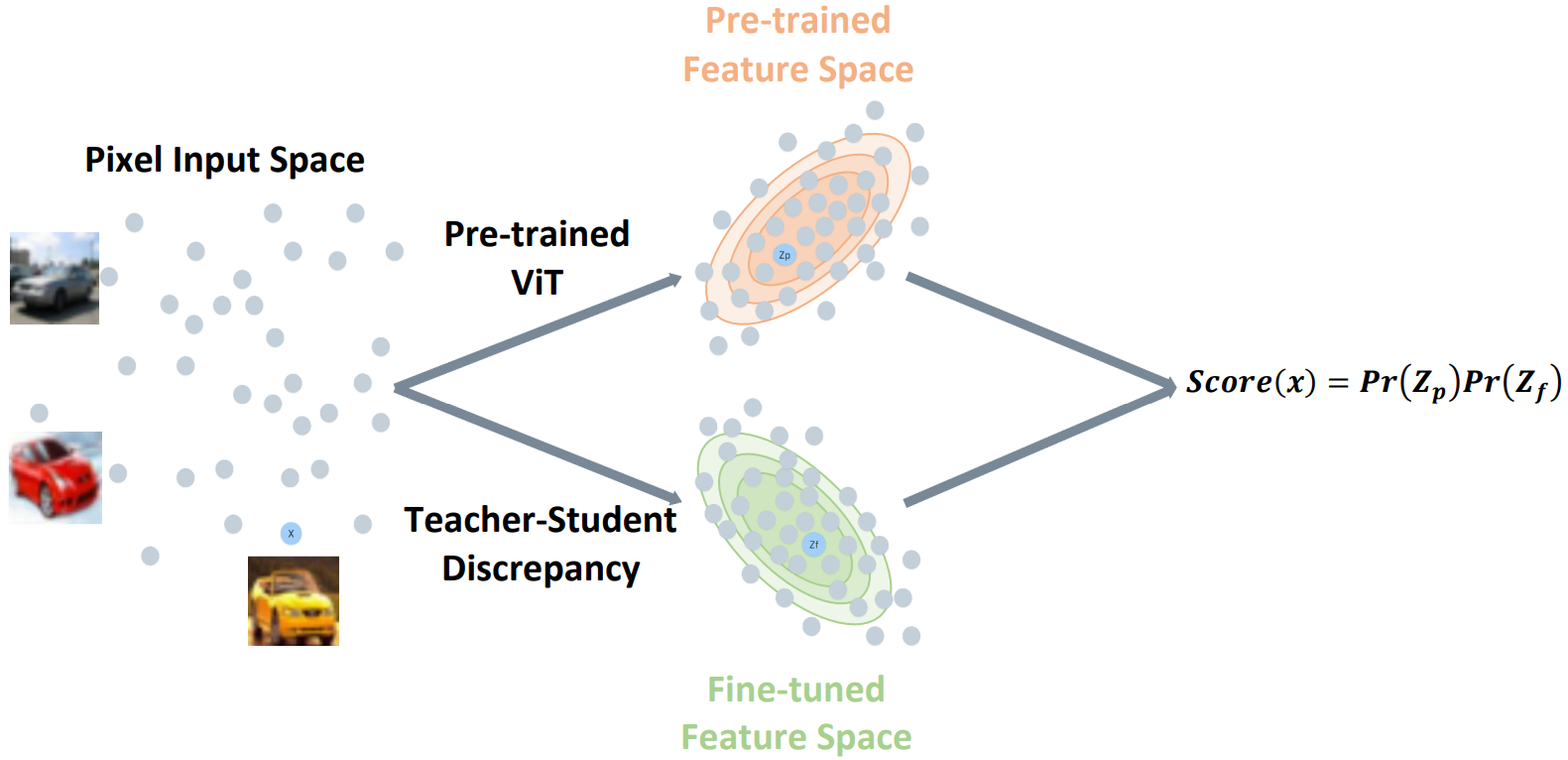}
\end{center}
  \caption{{\bf Transformaly at a glance:} The input data (left), ${\bf x}$, is mapped into both pre-trained features (top), ${\bf z_p}$, using pre-trained Visual Transformer (ViT) network, and fine-tuned features (bottom), ${\bf z_f}$, using teacher-student training to predict the output of different blocks in ViT. In each space we fit a Gaussian to the data. The likelihood of a query point is the product of its likelihood in both spaces. 
  Previous methods used either pre-trained features or fine-tuned features, but not both.
}
\label{fig:SchematicPlot}
\end{figure}
% Given a query point we compute its likelihood in each space and take Overall model architecture. Using the training normal data, we train randomly-initialized ViT blocks to predict the ten last blocks’ outputs of the pre-trained ViT. In addition, a Gaussian model was fitted on the pre-trained ViT penultimate layer's outputs, which was used as samples' embeddings. The final normality score is the product of the likelihood of two Gaussian models: one fitted using the sample pre-trained ViT embeddings (pre-trained features) while the other fitted using the discrepancy between the pre-trained ViT blocks' output and their corresponding student blocks' output (fine-tuned features).

Anomaly detection is a long-standing field of research that has many applications in computer vision. The realm of anomaly detection is broad and involves different types of problems.
% As in other cases, solutions to this problem are now dominated by deep learning techniques.

% The term "Anomaly Detection" has been used to describe several different tasks. 
In the case of multi-class classification, the term "Anomaly Detection" is often used to describe out-of-distribution (OOD) detection or novelty detection; where the task is to determine at inference time, if a test sample belongs to one of the classes the model was trained to classify or not. This task relies on what is known as the open-set assumption, where the class of a query sample can be outside the set of classes used in training.

% the realm of anomaly detection is broad and involves different types of problems. In general, 
Aside from OOD detection, one can consider two anomaly detection variants: (i) semantic anomaly detection, in which the normal and the abnormal samples differ in their semantic meaning; (ii) defect detection, in which the normal and the abnormal samples differ in their local appearance (i.e., defect), but are semantically identical. We consider the case of semantic anomaly detection.
% While there are papers that address both tasks with the same solution, these two problems have different characteristics and therefore, we believe, should be addressed differently.

% We consider the case of semantic anomaly detection. Specially, the case of one-class classification, where the goal is to train a model on data of one class, while at test time the model must determine whether a sample belongs to that class or not.

At a high level, anomaly detection involves the combination of representation and modelling. Solutions to the problem can be divided into three categories: the unsupervised approach, the self-supervised approach and the pretrained-based approach. 
%  An ideal representation will make it easy to model the normal data and separate it from the abnormal. 

Unsupervised methods use only normal data, without any form of labeling. These methods include methods for reconstructing the normal data~\cite{eskin2002geometric, xia2015learning, zenati2018efficient, li2018anomaly, deecke2018image}, estimating its density~\cite{glodek2013ensemble}, or concentrating it into one manifold~\cite{Ruff2018deep, scholkopf1999support, tax2004support}. 
% These method focus on the modeling part of the problem and do not attempt to find a new representation.
% \cite{ eskin2002geometric, xia2015learning, zenati2018efficient, li2018anomaly, deecke2018image, scholkopf1999support, tax2004support}. 

In the self-supervised approach, a model is trained on an auxiliary task. Hopefully, the model learns meaningful features that reflect the normal nature of the data. The construction of an auxiliary task that motivates the model to learn these relevant features is not trivial, and several suggestions have been made such as geometric transformations classification \cite{golan2018deep}, rotation classification \cite{hendrycks2019using}, puzzle-solving \cite{salehi2020puzzle} and CutPaste \cite{li2021cutpaste}.

Recently, significant progress has been made in the self-supervised domain, with the use of contrastive learning \cite{thulasidasan2020simple,he2020momentum,grill2020bootstrap}. Studies have shown that contrastive learning can be useful for semantic anomaly detection and produce good results \cite{tack2020csi, sohn2020learning}. 

The combination of feature extraction from a pre-trained model and simple scoring algorithm on top of it, is an effective approach for anomaly detection~\cite{bergman2020deep, hendrycks2019using, xiao2021we}. Bergman \textit{et al.} \cite{bergman2020deep} used a pre-trained ResNet model, and applied a $k$NN scoring method on the extracted features. That alone surpassed almost all unsupervised and self-supervised methods. Fine-tuning the model using either center loss or contrastive learning, leads to even better results~\cite{reiss2020panda,reiss2021mean}. 

On the downside, pre-trained features are agnostic to the normal data that is available at the training stage. It is a loss of valuable information, and we propose to address it by using teacher-student training. Specifically, our work combines both pre-trained features and teacher-student training. In both cases, we use a pre-trained Vision Transformer (ViT) network as our backbone~\cite{dosovitskiy2020image}. 

Teacher-student training was already used for pixel-precise anomaly segmentation in high resolution images~\cite{bergmann2020uninformed}. Their representation learns low level statistics that are suitable for defect detection tasks. They report results that are considerably sub-par for the case of semantic anomaly detection.

% Teacher-student training was already used for pixel-precise anomaly segmentation in high resolution images~\cite{bergmann2020uninformed}. Their representation learns low level statistics which makes suitable for Defect Detection tasks where semantics is irrelevant. However, this method does not preform competitively well using semantic anomaly detection \footnote{more than 15\% degradation from SOTA on cifar10, as reported in the paper}.

We, on the other hand, focus on semantic anomaly detection, where semantic representation is crucial. In contrast to~\cite{bergmann2020uninformed}, we utilize not only the teacher-student discrepancy representation, but also the raw pre-trained embedding, achieving SOTA results in detecting semantic anomalies. 

% This pre-trained embedding space has been empirically demonstrated to be semantically separable (Figure \ref{fig:cifar10_tsne} in the supplementary). Thus, it facilitates anomaly detection, especially if the student blocks somehow generalize their teacher's functionality to some abnormal samples. The student blocks in Transformaly mimic several teacher blocks, using the pre-trained intermediate outputs instead of just the ultimate output. 

We modify the standard teacher-student setting by using blocks instead of the entire network. Specifically, we use the Vision-Transformer architecture (ViT) and exploit the nature of its block structure. We construct a student block that corresponds to each block of the teacher backbone. At train time, each student block is trained to mimic its corresponding teacher block. The student blocks are trained independently, and are exposed only to normal samples. At test time, each sample is represented by a vector of the differences between the teachers' outputs and the students' outputs. We term this representation teacher-student discrepancy. 
% Intuitively, small differences indicate normal examples and large values indicate abnormal samples. These differences form the fine-tuned features of the data. We term this representation teacher-student discrepancy. 

Figure~\ref{fig:SchematicPlot} gives a high-level overview of our approach. Each sample is mapped to two different feature spaces: one created by a pre-trained ViT network (the agnostic features) and another created by the discrepancy between student and teacher blocks (fine-tuned features). 

The likelihood of a sample is taken to be the product of its likelihood in both feature spaces. To model the likelihoods we experimented with several options that include $k$NN, a single Gaussian, and a Gaussian Mixture Model. 

We evaluate our method on several data sets, and find that in most cases this combined representation outperforms existing state-of-the-art methods. The main contributions of this paper are:

\begin{itemize}[noitemsep]
\item Transformaly - a first use of dual feature representation for anomaly detection: agnostic and fine-tuned.
\item A novel use of teacher-student differences using Visual Transformers (ViT) for semantic anomaly detection.
\item State of the art results on multiple datasets and multiple settings: Cutting the error by $40-65\%$ on competitive benchmarks.
\item We are the \emph{first} to report a hard-to-detect failure in the common unimodal setting, which we called \emph{"Pretraining Confusion"}, justify our additional multimodel evaluation process.
\end{itemize}

\section{Background}

The term “Anomaly Detection” encapsulate several different tasks in the literature. It has been used to describe Out-of-Distribution Detection, Defect Detection, and Semantic Anomaly Detection (the topic of this work). We will briefly cover these tasks here.

% Aside from OOD detection, the realm of anomaly detection is very broad, comprehensive, and involves different types of problems. In general, there are two main tasks in the anomaly detection domain: i) semantic anomaly detection- the normal and the abnormal samples differ in their semantic meaning. ii) defect detection- the normal and the abnormal samples differ in a local region (defect), but they are semantically identical. While there are papers that address both tasks with the same solution, these two problems have different characteristics and therefore, we believe, should be addressed differently.

\paragraph{\bf Out Of Distribution Detection:}
Out-of-distribution (OOD) detection, also known as novelty detection, considers the case of multi-class classification. In this task, in addition to training an accurate model, we would also like to detect when it encounters a sample that does not belong to any of the known classes.

Several approaches took advantage of the power of supervised multi-class classification. For example, the model predictions of out-of-distribution samples have lower values than those of in-distribution samples, making anomaly detection possible \cite{hendrycks2016baseline}. Another notable approach has shown that the gap between temperature-scaled softmax scores of a sample and a perturbed version of it can be measured and used as a normality score \cite{liang2017enhancing}. It turns out that a larger gap is apparent in anomalous samples than in normal ones.
% It has been shown, that one can measure the gap between temperature-scaled softmax scores of a sample, and a perturbated version of it, and use this gap as a normality score \cite{liang2017enhancing}. It turns out that a larger gap is apparent in anomalous samples than in normal ones. 
% Moreover, using temperature scaling on the softmax scores, as well as small perturbations to the input samples, will increase the gap in softmax scores between in-distribution and out-of-distribution observations, facilitating detection \cite{liang2017enhancing}. 

In addition, using a small dataset of possible anomalies can boost the detection performance \cite{hendrycks2018deep}. More recent approaches use an ensemble of models, pre-trained transformers, or an extra abstention class for detecting out-of-distribution samples \cite{vyas2018out, hendrycks2020pretrained, thulasidasan2020simple}. %While this topic is interesting and highly motivated, we will focus on a  anomaly detection task instead.

\paragraph{\bf Defect Detection:}
In defect detection the normal and the abnormal samples differ in local appearance, but are semantically identical. For example, defects in printed circuits, cables or medicinal pills.

Researchers proposed several approaches for defect detection \cite{venkataramanan2020attention, bergmann2019mvtec, bergmann2020uninformed}. Two recent papers use the Vision Transformer (ViT) architecture; Mishra {\em et al.}~\cite{mishra2021vt} suggest using encoder-decoder architecture in order to reconstruct the normal data. Pirany and Chai~\cite{pirnay2021inpainting} proposed to train ViT using the auxiliary task of patch-inpainting. At inference time both methods use the discrepancy between the input image patches and the reconstructed image patches as an indication of possible defects. 

% Researchers proposed several approaches for defect detection \cite{venkataramanan2020attention, bergmann2019mvtec, bergmann2020uninformed}. Two recent papers use the Vision Transformer (ViT) architecture; Mishra {\em et al.}~\cite{mishra2021vt} suggest using encoder-decoder architecture in order to reconstruct the normal data. The authors used ViT as their encoder architecture and applied a Gaussian mixture density network on its outputs as well. Pirany and Chai~\cite{pirnay2021inpainting} proposed to train vision transformer using the auxiliary task of patch-inpainting. At inference time both of the methods use the discrepancy between the input image patches and the reconstructed image patches as an indication of possible defects. 

Neither of these solutions includes a pre-training phase for the ViT. Both are best suited to detecting local defects reflected in patches, not semantic anomalies.

Bergmann \textit{et al.} \cite{bergmann2020uninformed} suggested a teacher-student architecture for defect detection. The teacher model is based on a ResNet model pre-trained on a large dataset of patches from natural images. Then, an ensemble of student networks is trained on anomaly-free training data using regression loss with the teacher’s ultimate outputs.

This patch-based method is suited for defect detection, where the anomaly is in appearance and not semantic. Furthermore, this method uses only the last layer outputs for the teacher-student mechanism, and does not utilize the semantic embedding of the pre-trained network. Transformaly, on the other hand, takes advantage of different blocks in the model, by training the student blocks using intermediate outputs. Additionally, the semantically pre-trained embedding is used along with fine-tuned embedding to obtain SOTA results.

\begin{figure*}
\begin{center}
\includegraphics[width=1\linewidth]{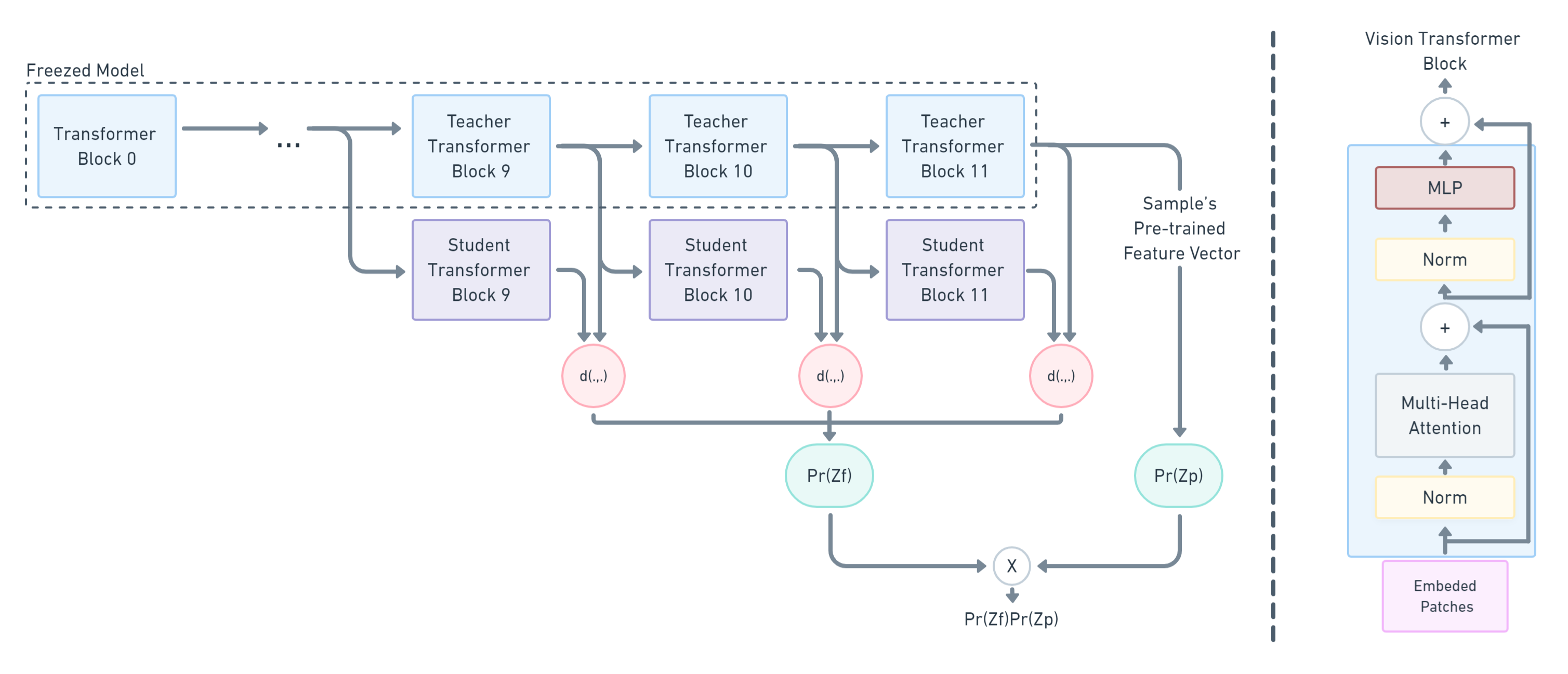}

\end{center}
  \caption{{\bf Transformaly architecture:} We use ViT to produce pre-trained features (top part). The same ViT network is used as a teacher network to train a student network (with the same architecture) on the normal training data. The discrepancy between student and teacher networks forms the fine-tuned features. The data in each space is fitted with a Gaussian, and the final normality score is the product of the likelihood of the two Gaussian models.}
\label{fig:ModelArchitecture}
\end{figure*}

\paragraph{\bf Semantic Anomaly Detection:}
We can classify semantic anomaly detection solutions into three categories:  Unsupervised Learning, self-supervised, and pretrained-based approaches.

Unsupervised anomaly detection solutions fall into three approaches:  reconstruction-based, density estimation, and one-class classifiers. 

Reconstruction-based methods attempt to capture the main characteristics of a normal training set by measuring reconstruction success. By assuming that only normal data will reconstruct well at test time, these methods attempt to detect anomalies.  Previous papers have suggested $k$ nearest neighbors ($k$-NN) \cite{eskin2002geometric}, autoencoder \cite{xia2015learning} and GANs \cite{zenati2018efficient, li2018anomaly, deecke2018image} in order to reconstruct and classify the samples. 
% In order to score samples, these methods use a reconstruction metric, such as $L_2$, as the normality score.

Density-based methods estimate the density of the normal data. These methods predict the samples' likelihood as their normality scores. Previous papers suggested parametric density estimation, such as mixture of Gaussians (GMM) \cite{glodek2013ensemble}, and nonparametric, such as $k$-NN \cite{peterson2009k}.  

One-class classification methods map normal data to a manifold, leaving the abnormal samples outside. A few modifications have been made to SVM to adjust it for this purpose, with training only on one class \cite{Ruff2018deep, scholkopf1999support, tax2004support}.

Self-supervised methods use auxiliary tasks in order to learn relevant features of the normal data. These methods train a neural network to solve an unrelated task, using just the normal training data. At inference time,  the model's auxiliary task performance on the test set is considered as its normality score.

A number of papers have proposed applying transformations to the normal data and predicting which transformations have been applied. Predicting predefined geometric transformations \cite{golan2018deep}, rotation \cite{hendrycks2019using}, puzzle-solving \cite{salehi2020puzzle} and CutPaste \cite{li2021cutpaste} are a few of the auxiliary tasks that have been suggested. In addition, (random) general transformations can be applied not only to images but also to tabular data, enabling anomalies to be detected in this domain as well \cite{bergman2020classification}. 

% Developing an auxiliary task that produces features that can be used to detect anomalies is the key challenge of this approach. 
Recent papers demonstrate the effectiveness of contrastive learning as a self-supervised method for learning visual representations, achieving SOTA results \cite{chen2020simple,he2020momentum,grill2020bootstrap}. A contrastive learning approach, such as SimCLR, produces uniformly distributed outputs, making anomalies difficult to spot \cite{chen2020simple, sohn2020learning}. Using contrasting shifted instances, along with rotation or transformation prediction, managed to surpass this challenge and achieved high performance \cite{tack2020csi}. Additionally, training a feature extractor using shifted contrastive learning and applying one class classification or kernel density estimation has been shown to be effective for anomaly detection \cite{sohn2020learning}.  

Pretrained-based methods use backbones that are trained on large datasets, such as ImageNet, to extract features \cite{deng2009imagenet}. While in the past obtaining pre-trained models was a limitation; these days they are readily available and commonly used across many domains. These pre-trained models produce separable semantic embeddings and, as a result, enable the detection of anomalies by using simple scoring methods such as $k$-NN or Gaussian Mixture Model \cite{bergman2020deep, xiao2021we}. 

Surprisingly, the embeddings produced by these algorithms lead to good results also on datasets that are drastically different from the pretraining one. 
A follow-up paper improves the pre-training model detection performance by fine tuning it using center loss \cite{reiss2020panda}. Recent publication has suggested to fine-tuned the pre-trained network using an additional dataset as outlier exposure, to further boost the results \cite{deecke2021transfer}.

\section{Method}

We exploit the power of pre-trained ViT by constructing two features spaces for each sample; the pre-trained and the fine-tuned. A sample ${\bf x}$ is embedded into a pre-trained feature vector ${\bf z_p}$ and a fine-tuned feature vector ${\bf z_f}$. 

% The pre-trained vector ${\bf z_p}$ is obtained by passing the input ${\bf x}$ through a pre-trained Visual Transformer (ViT) network. This produces an embedding that is agnostic to the actual normal and abnormal data at hand. The fine tuned feature vector ${\bf z_f}$ depends on the actual normal data and is obtained by teacher-student training. The combination of both embedding squeezes as much information from the pre-trained ViT model and the normal data as possible.

% \paragraph{Pre-trained Features} \label{SubSectionGM}
% We set ${\bf e_p} = ViT({\bf x})$. Then, we whiten and project the data to a lower dimension, noted by ${\bf z_p}$, and fit a Gaussian to it. At inference time, each sample is scored by its log probability using the fitted model. We note this score by $GM({\bf z_p})$. The log probability of normal samples is assumed to be higher than that of abnormal samples.

\paragraph{Pre-trained Features} \label{SubSectionGM}
The pre-trained vector ${\bf z_p}$ is obtained by passing the input ${\bf x}$ through a pre-trained ViT network. This produces an embedding that is agnostic to the actual normal and abnormal data at hand.

We set ${\bf z_p} = \mbox{ViT}({\bf x})$ and fit a Gaussian to it. At inference time, each sample is scored according its log probability as induced by the fitted model. Normal samples are assumed to have higher probability than abnormal samples.

\paragraph{Fine-tuned features} \label{SubSectionBOP}
% Blocks’ outputs prediction (BoP) is inspired by the knowledge distillation domain \cite{wang2021knowledge}. 

The fine-tune feature embedding is inspired by the knowledge distillation domain \cite{wang2021knowledge}. It is calculate as the difference between the output of a teacher and student ViT blocks. Specifically, we train the student block only on normal data, such that it produces an output that is similar to the teacher output. The output of the student block is expected to be quite different when the data is abnormal.

We followed this process for $m$ different teacher-student blocks. We train each student block $h'_j$ independently to mimic $h_j$ using an $MSE$ loss:
\begin{equation}
    {\cal L} = \frac{1}{n} \sum_{i=1}^{n}|h_j({\bf x_i})-h'_j({\bf x_i})|^2_2
\end{equation}
At inference time, sample ${\bf x}$ is represented with:
\begin{equation}
    {\bf z_f} = [{\bf z_f}^{(0)},...,{\bf z_f}^{(m-1)}]
\end{equation}
where $m$ is the number of blocks in the ViT network, and  ${\bf z_f}^{(j)} = || h_j({\bf x_i})-h'_j({\bf x_i}) ||^2_2$ is the difference between the $j$-th teacher block $h_j$ and the $j$-th student block $h'_j$. We typically use $m=10$ to model the last 10 blocks in the ViT network. We empirically observed that using the first two blocks does not improve the model's performance. The first two layers may have learned low-level features that appear in both normal and abnormal samples. As such, these features are useless for detecting semantic anomalies.

\paragraph{Final Scoring method} \label{SubSectionScoringMethod}
We fit two Gaussians to both the pre-trained embedding ${\bf z_p}$ and the fine-tuned embedding ${\bf z_f}$.
\begin{equation}
    Pr({\bf z_p}) = {\cal N}( {\bf z_p} | {\bf \mu_p}, {\bf \Sigma_p})
\end{equation}
\begin{equation}
    Pr({\bf z_f}) = {\cal N}( {\bf z_f} | {\bf \mu_f}, {\bf \Sigma_f})
\end{equation}
where ${\bf \mu_{p}},{\bf \Sigma_{p}}$ are the mean and covariance of the pre-trained embeddings, and ${\bf \mu_{f}},{\bf \Sigma_{f}}$ are the mean and covariance of the fine-tuned embeddings. The final score of sample ${\bf x}$ is simply the product of the two (or the sum of their log):
\begin{equation}
    \mbox{score}({\bf x}) = Pr({\bf z_p})Pr({\bf z_f})
\end{equation}
Despite the fact that the likelihoods of the two Gaussians are not independent, we chose their product as our normality scoring method \footnote{This choice is guided by simplicity. We were motivated to find a simple operation that flips the verdict only when one score is larger/lower than the other by orders of magnitude.}. This method and scoring procedure is used in all the experiments in the next section, unless otherwise stated.

\section{Experiments}
%-------------------------------------------------------------------------
%-------------------------------------------------------------------------

We describe implementation details, the benchmark settings and datasets that we used, and present the results of the experiments in the following sub-sections. 
% Each of the scores presented in tables 1-4 is the average of the AUROC scores across classes in the dataset.

\subsection{Implementation Details}
We use a PyTorch implementation of ViT, trained on ImageNet-21k and fine-tuned on ImageNet-1k \cite{melas-kyriazi_vit_2021, ridnik2021imagenet}. ViT has $12$ heads, $16 \times 16$ input patch size, dropout rate of $0.1$ and its penultimate layer outputs $768$-dimensional vectors, which form the pre-trained features of our method. All the input images are normalized according to the pre-training phase of the ViT. Unless otherwise specified, the fine-tune features are $10D$ vectors that are taken to be the result of applying teacher-student training to the last ten blocks of ViT. In each feature space we model the data with a single Gaussian using its mean and full covariance. Since pre-trained features live in a $768D$ space, we first whiten and reduce the dimensionality of these features by keeping the number of components that explain $90\%$ of the data variance (Typically, this results in vector a of 300 dimensions).

% \begin{figure}[h!]
% \begin{center}
% \includegraphics[width=1\linewidth]{latex/figures/unimodal_vs_multimodal.png}

% \end{center}
%   \caption{{\bf Unimodal Vs. multimodal:} A toy example demonstrating the difference between the two settings. Each shape represents a different class in the dataset. In a unimodal setting only squares are considered normal. In a multimodal setting all classes except squares are considered normal samples. The situation is not symmetric and affects AUROC scores. See text for details.}
% \label{fig:unimodal_vs_multimodal}
% \end{figure}

\subsection{Datasets}

\begin{figure}[t]
\begin{center}
\includegraphics[width=1\linewidth]{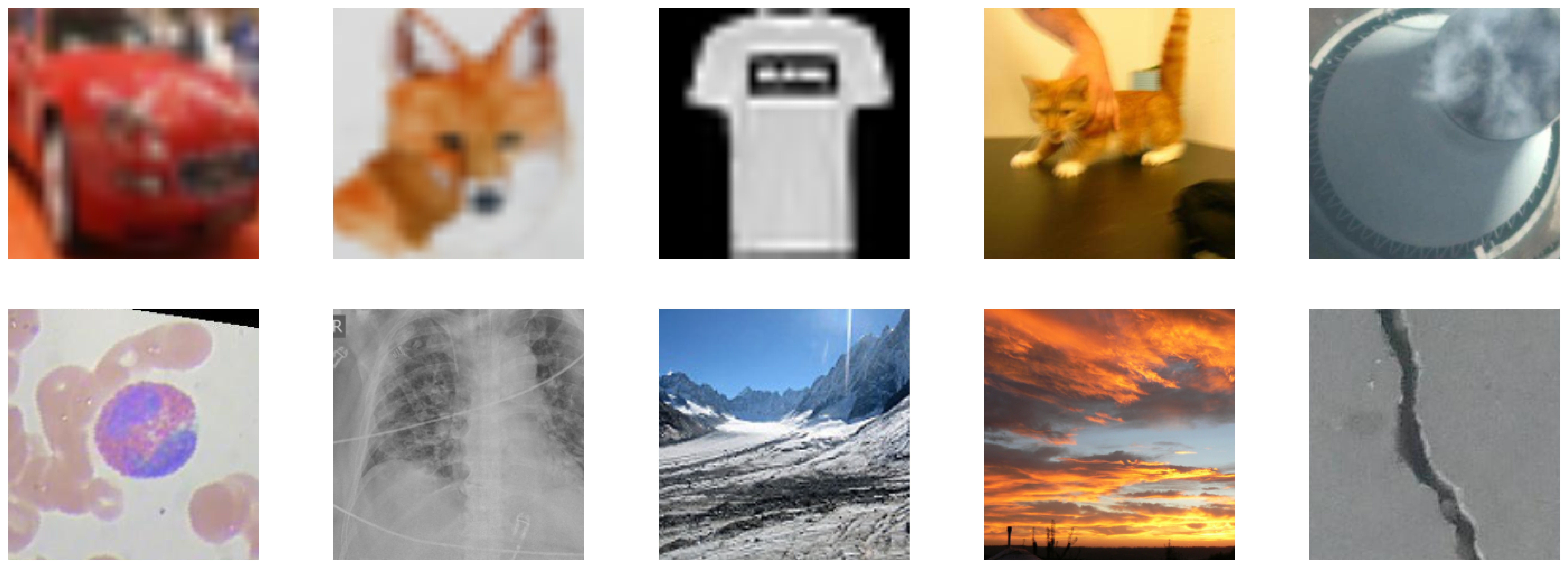}
\end{center}
  \caption{ \bf{A representative image for each dataset we used, from left to right, top to bottom:} Cifar10, Cifar100, Fashion MNIST, DogsVsCats, DIOR, Blood Cells, Covid19, View Recognition, Weather Recognition, Concrete Crack Classification.}
\label{fig:datasets_exampler}
\end{figure}

Transormaly is evaluated on commonly used datasets: Cifar10, Cifar100, Fashion MNIST, and Cats vs Dogs. We evaluated Transformaly's robustness using additional datasets: aerial images (Dior), blood cell images (Blood Cells), X-ray images of Covid19 patients (Covid19), natural scenes images (View Recognition), weather image (Weather Recognition) and images of plain and cracked concrete (Concrete Crack Classification). We show a representative image from each dataset in Figure~\ref{fig:datasets_exampler}. As can be seen, the datasets are quite diverse. Please refer to the supplementary material for more details on the datasets used.% (Section \ref{datasets_details}). 

\begin{table*}[t]
\centering
\begin{tabular}{|l|l|l|l|l|l|l|l|l|}
\hline
Dataset    &
\multicolumn{2}{c|}{Unsupervised} &
\multicolumn{2}{c|}{Self-Supervised} &
\multicolumn{4}{c|}{Pretrained} \\
\hline &
\multicolumn{1}{l|}{OC-SVM} &
\multicolumn{1}{l|}{DeepSVDD} &
\multicolumn{1}{l|}{MHRot} &
\multicolumn{1}{l|}{CSI} &
\multicolumn{1}{l|}{DN2} &
\multicolumn{1}{l|}{PANDA} &
\multicolumn{1}{l|}{MSAD} &
% \multicolumn{1}{l}{Pretrain ViT + BOP} \\
\multicolumn{1}{l|}{Ours}\\
\hline
CIFAR10    &
\multicolumn{1}{|c|}{64.7} &
\multicolumn{1}{c|}{64.8} &
\multicolumn{1}{c|}{90.1} &
\multicolumn{1}{c|}{ 94.3} &
\multicolumn{1}{c|}{ 92.5} &
\multicolumn{1}{c|}{96.2} &
\multicolumn{1}{c|}{\underline{97.2}} &
\multicolumn{1}{c|}{\textbf{98.31}} \cr 
\cline{1-1}
CIFAR100   &
\multicolumn{1}{|c|}{62.6} &
\multicolumn{1}{c|}{67.0} &
\multicolumn{1}{c|}{80.1} &
\multicolumn{1}{c|}{ 89.6} &
\multicolumn{1}{c|}{94.1} &
\multicolumn{1}{c|}{94.1} &
\multicolumn{1}{c|}{\underline{96.4}} &
\multicolumn{1}{c|}{\textbf{97.34}} \cr 
\cline{1-1}
FMNIST     &
\multicolumn{1}{|c|}{92.8} &
\multicolumn{1}{c|}{84.8} &
\multicolumn{1}{c|}{93.2} &
\multicolumn{1}{c|}{-} &
\multicolumn{1}{c|}{ \underline{94.5}} &
\multicolumn{1}{c|}{\textbf{95.6}} &
\multicolumn{1}{c|}{94.21} &
\multicolumn{1}{c|}{94.43} \cr 
\cline{1-1}
CatsVsDogs &
\multicolumn{1}{|c|}{51.7} &
\multicolumn{1}{c|}{50.5} &
\multicolumn{1}{c|}{86.0} &
\multicolumn{1}{c|}{$86.3^{\natural}$} &
\multicolumn{1}{c|}{96.0} &
\multicolumn{1}{c|}{97.3} &
\multicolumn{1}{c|}{\underline{99.3}} &
\multicolumn{1}{c|}{\textbf{99.52}} \cr
\cline{1-1}
DIOR &
\multicolumn{1}{|c|}{$70.7^{\sharp}$} &
\multicolumn{1}{c|}{$70.0^{\sharp}$} &
\multicolumn{1}{c|}{$73.3^{\sharp}$} &
\multicolumn{1}{c|}{$78.5^{\natural}$} &
\multicolumn{1}{c|}{ 92.2} &
\multicolumn{1}{c|}{94.3} &
\multicolumn{1}{c|}{\underline{97.2}} &
\multicolumn{1}{c|}{\textbf{98.08}} \cr 
\hline
\end{tabular}
\caption{{\bf AUROC scores of the unimodal setting:} We compare our method (rightmost column) against the alternatives. We outperform all methods on all datasets, except for FMNIST (All values, except for our method, are taken from~\cite{reiss2020panda}, except the value $94.21$ of MSAD~\cite{reiss2021mean} on FMNIST that was computed by us using the official code released by the authors). $\sharp$ taken from \cite{reiss2020panda}, $\natural$ taken from \cite{reiss2021mean}.
}
\label{table:unimodal}
\end{table*}

\subsection{Benchmark Settings}

We examine Transformaly in the unimodal and multimodal settings. In the unimodal case, one semantic class is randomly chosen as normal while the rest of the classes are treated as abnormal. During training, the model is only exposed to the normal class’ samples. At inference time, all samples of the test set are evaluated, while samples that are not from the normal class are considered anomalous. In the multimodal case the roles are reversed, and all classes but one are considered normal.

% We evaluate both settings because, while the unimodal setting is common in the literature, it does not adequately reflect real-life scenarios, 
We evaluate both settings because the unimodal setting, while common in the literature, does not adequately reflect all real-life scenarios; where normal data might contain multiple semantic classes. Moreover, unimodal setting leads to a peculiar evaluation process, where we have much more anomalies than normal samples (nine and nineteen times more for Cifar10 and Cifar100, respectively). This is not in line with the common anomaly detection use-case where the normal samples are the majority of the data that the model encounters and the anomalies are the rare events.

% On top of that, the two settings are not symmetric, as is illustrated in Figure~\ref{fig:unimodal_vs_multimodal}. It shows a toy example of four semantically different classes (triangles, diamonds, circles, and squares). The triangles and diamonds are nicely separated, while the squares and circles are confused. 

% Consider the unimodal case, where only the blue squares are available as normal samples during training. In this case, at test time only the red circles will be confused as normal instead of abnormal. The red triangles and diamonds will be correctly classified as abnormal. The algorithm misses {\em some} of the abnormalities.

% The situation is reversed in the multimodal case. Assume now that all red samples (triangles, diamonds, and circles) are normal. At test time, all the abnormal blue squares will be classified as normal. The algorithm misses {\em all} the abnormalities.

It should be noted that this problem was recently discussed in the context of OOD detection by Courville and Ahmed who criticized the standard benchmark in OOD detection \cite{ahmed2020detecting}.  We adopt their proposed multimodal paradigm for anomaly detection as complementary evaluation process to the common unimodal setting. %We believe that the two settings are necessary in order to fully evaluate any anomaly detection solution.

% The OOD detection common benchmark is to use one dataset for the Multiclass classification task, and a second dataset for the evaluation of the OOD detector. Courville and Ahmed claimed that it is ill-motivated and the OOD detection can be explained by low-level image statistics, or by underlying biases of the two datasets. The authors suggest using a single dataset for both the Multiclass classification task and the OOD detection. Their suggestions was to treat one class as an out-of-distribution and all other classes as in-distributions, which will be used for classification. 

% We adopt their proposed paradigm for anomaly detection as complementary evaluation process to the common unimodal setting. In this setting, we treat all classes as normal but one, which we treat as abnormal. During the training process, we do not use the labels of the normal samples but simply treat them as a single multimodal normal class. We believe that the two settings are necessary in order to fully evaluate any anomaly detection solution. This is why we show results in the multimodal setting as well. 

\subsection{Results}

Table~\ref{table:unimodal} shows results of the common unimodal setting, in which one class is considered normal while all other classes are abnormal. A threshold-free area under the receiver operating (AUROC) characteristic curve is used to evaluate the models. We report the performance of our method and compare it  to unsupervised, self-supervised and pre-trained based methods. Each of the scores presented in following tables is the average of the AUROC scores across all classes in each dataset. One can observe that our method outperforms all other methods on all datasets, except for Fashion MNIST dataset.

\begin{table}[]
\small
\centering
\begin{tabular}{|l|l|l|l|l|}
\hline
Dataset    &
\multicolumn{1}{l|}{DeepSVDD} &
\multicolumn{1}{l|}{DN2} &
\multicolumn{1}{l|}{PANDA} &
\multicolumn{1}{l|}{Ours}\\
\hline
Blood Cells    &
\multicolumn{1}{c|}{52.28} &
\multicolumn{1}{c|}{54.91} &
\multicolumn{1}{c|}{\underline{56.25}} &
\multicolumn{1}{c|}{\textbf{74.85}} \cr 
\cline{1-1}
Covid19   &
\multicolumn{1}{c|}{97.32} &
\multicolumn{1}{c|}{97.63} &
\multicolumn{1}{c|}{\textbf{99.33}} & 
\multicolumn{1}{c|}{\underline{98.87}} \cr 
\cline{1-1}
% Weather Recognition     &
Weather     &
\multicolumn{1}{c|}{73.55} &
\multicolumn{1}{c|}{80.05} &
\multicolumn{1}{c|}{\underline{81.53}} &
\multicolumn{1}{c|}{\textbf{94.32}} \cr 
\cline{1-1}

% View Recognition &
View &
\multicolumn{1}{c|}{60.31} &
\multicolumn{1}{c|}{90.86} &
\multicolumn{1}{c|}{\underline{93.63}} &
\multicolumn{1}{c|}{\textbf{95.80}} \cr
\cline{1-1}

% \thead{Concrete Crack \\ Classification }&
Concrete &
\multicolumn{1}{c|}{92.27} &
\multicolumn{1}{c|}{\underline{99.81}} &
\multicolumn{1}{c|}{\textbf{99.93}} &
\multicolumn{1}{c|}{99.77} \cr 
\hline
\end{tabular}
\caption{{\bf AUROC scores of the unimodal setting on various datasets:} We compare our method against some of the alternatives. We outperform other methods on most datasets, sometimes by a large margin (over 18\% and 12\% on "Blood Cells" and "Weather Recognition" respectively), while we underperform only slightly on "Concrete Crack Classification" and "Covid19", where we come in second.}
\label{table:additional_dataset}
\end{table}

\noindent
\begin{table*}[h]
\small
\centering
\begin{tabular}{|l!{\vrule width 1.0pt}l|l!{\vrule width 1.0pt}l|l|l|l|l|}
\hline
Dataset    &
\multicolumn{1}{c!{\vrule width 1.0pt}}{Pre-trained} &
\multicolumn{3}{c!{\vrule width 1.0pt}}{Fine-tuned} &
\multicolumn{3}{c|}{Full Model} \\
% \hline & \multicolumn{1}{c!{\vrule width 0.8pt}}{Gaussian} &
\hline & \multicolumn{1}{c!{\vrule width 1.0pt}}{} & 
\multicolumn{1}{p{1.5cm}|}{\centering {Last 1 block}} & \multicolumn{1}{p{1.5cm}|}{\centering {Last 3 blocks}} &
\multicolumn{1}{p{1.5cm}!{\vrule width 1.0pt}}{\centering {Last 10 blocks}} &
\multicolumn{1}{p{1.5cm}|}{\centering {Last 1 block}} &
\multicolumn{1}{p{1.5cm}|}{\centering {Last 3 blocks}} &
\multicolumn{1}{p{1.5cm}|}{\centering {Last 10 blocks}} \\
\hline
CIFAR10    &
\multicolumn{1}{c!{\vrule width 1.0pt}}{97.81} &
\multicolumn{1}{c|}{95.02} & 
\multicolumn{1}{c|}{97.18} &
\multicolumn{1}{c!{\vrule width 1.0pt}}{96.63} &
\multicolumn{1}{c|}{\underline{98.27}} &
\multicolumn{1}{c|}{98.26} &
\multicolumn{1}{c|}{\textbf{98.31}} \cr 
\cline{1-1}
CIFAR100   &
\multicolumn{1}{c!{\vrule width 1.0pt}}{96.21} &
\multicolumn{1}{c|}{90.71} &
\multicolumn{1}{c|}{94.79} &
\multicolumn{1}{c!{\vrule width 1.0pt}}{95.16} &
\multicolumn{1}{c|}{96.90} &
\multicolumn{1}{c|}{\underline{96.99}} &
\multicolumn{1}{c|}{\textbf{97.34}} \cr 
\cline{1-1}
FMNIST     &
\multicolumn{1}{c!{\vrule width 1.0pt}}{93.94} &
\multicolumn{1}{c|}{88.41} &
\multicolumn{1}{c|}{92.39} &
\multicolumn{1}{c!{\vrule width 1.0pt}}{\underline{94.14}} &
\multicolumn{1}{c|}{94.00} &
\multicolumn{1}{c|}{94.07} &
\multicolumn{1}{c|}{\textbf{94.43}} \cr 
\cline{1-1}
CatsVsDogs     &
\multicolumn{1}{c!{\vrule width 1.0pt}}{\underline{99.60}} &
\multicolumn{1}{c|}{98.30} &
\multicolumn{1}{c|}{97.45} &
\multicolumn{1}{c!{\vrule width 1.0pt}}{96.47} &
\multicolumn{1}{c|}{\textbf{99.66}} &
\multicolumn{1}{c|}{99.58} &
\multicolumn{1}{c|}{99.52} \cr 
\cline{1-1}
DIOR     &
\multicolumn{1}{c!{\vrule width 1.0pt}}{93.97} &
\multicolumn{1}{c|}{94.72} &
\multicolumn{1}{c|}{95.26} &
\multicolumn{1}{c!{\vrule width 1.0pt}}{\textbf{98.59}} &
\multicolumn{1}{c|}{95.22} &
\multicolumn{1}{c|}{95.31} &
\multicolumn{1}{c|}{\underline{98.08}} \cr
\cline{1-1}
Blood Cells     &
\multicolumn{1}{c!{\vrule width 1.0pt}}{72.19} &
\multicolumn{1}{c|}{\underline{74.80}} &
\multicolumn{1}{c|}{75.43} &
\multicolumn{1}{c!{\vrule width 1.0pt}}{73.58} &
\multicolumn{1}{c|}{73.15} &
\multicolumn{1}{c|}{74.41} &
\multicolumn{1}{c|}{\textbf{74.85}} \cr 
\cline{1-1}
Covid19     &
\multicolumn{1}{c!{\vrule width 1.0pt}}{97.06} &
\multicolumn{1}{c|}{92.67} &
\multicolumn{1}{c|}{96.24} &
\multicolumn{1}{c!{\vrule width 1.0pt}}{\textbf{99.40}} &
\multicolumn{1}{c|}{97.13} &
\multicolumn{1}{c|}{97.37} &
\multicolumn{1}{c|}{\underline{98.87}} \cr 
\cline{1-1}
% Weather Recognition     &
Weather  &
\multicolumn{1}{c!{\vrule width 1.0pt}}{81.06} &
\multicolumn{1}{c|}{80.62} &
\multicolumn{1}{c|}{80.32} &
\multicolumn{1}{c!{\vrule width 1.0pt}}{\underline{94.26}} &
\multicolumn{1}{c|}{82.17} &
\multicolumn{1}{c|}{82.11} &
\multicolumn{1}{c|}{\textbf{94.32}} \cr 
\cline{1-1}
% View Classification &
View &
\multicolumn{1}{c!{\vrule width 1.0pt}}{95.48} &
\multicolumn{1}{c|}{94.40} &
\multicolumn{1}{c|}{94.41} &
\multicolumn{1}{c!{\vrule width 1.0pt}}{94.68} &
\multicolumn{1}{c|}{\underline{95.56}} &
\multicolumn{1}{c|}{95.54} &
\multicolumn{1}{c|}{\textbf{95.80}}\cr 
\cline{1-1}
% \thead{Concrete Crack \\ Classification} &
Concrete &
\multicolumn{1}{c!{\vrule width 1.0pt}}{99.72} &
\multicolumn{1}{c|}{99.47} &
\multicolumn{1}{c|}{99.13} &
\multicolumn{1}{c!{\vrule width 1.0pt}}{99.41} &
\multicolumn{1}{c|}{\underline{99.75}} &
\multicolumn{1}{c|}{99.74} &
\multicolumn{1}{c|}{\textbf{99.77}} \cr 
\hline
\end{tabular}
\caption{{\bf AUROC scores of pre-trained and fine-tuned features in the unimodal setting:} We compare the performance of pre-trained (leftmost column), fine-tuned features with different number of teacher-student blocks (middle 3 column), and the combined effect of pre-trained and various fine-tuned features (rightmost three columns).  In all cases, the data in each feature space is modeled with a single Gaussian. Fine-tuned features (generated by Teacher-Student discrepancy) provide, by themselves, satisfactory results. Fine-tuned features using 
only to the last ViT block ("Fine-Tuned, Last 1 block") represent a lightweight variant of our method that gives good results. On almost all datasets, fine-tuned features boost the performance of pre-trained features, sometimes by a large margin (i.e., weather recognition).}
\label{table:unimodal_bop_ablation}
\end{table*}

\begin{table}[]
\small
\centering
\begin{tabular}{|l|l|l|l|l|l|}
\hline
Dataset    &
\multicolumn{1}{l|}{\thead{$k$-NN \\ $k=2$ }} &
\multicolumn{1}{l|}{\thead{$k$-NN \\ $k=5$ }} &
% \multicolumn{1}{l|}{\thead{$k$-NN \\ $k=10$ }} &
\multicolumn{1}{l|}{\thead{GMM \\ $n=1$ }} &
\multicolumn{1}{l|}{\thead{GMM \\ $n=5$ }} &
% \multicolumn{1}{l|}{\thead{GMM \\ $n=10$ }} &
\multicolumn{1}{l|}{\thead{GMM \\ $n=20$ }} \\
\hline
CIFAR10    &
\multicolumn{1}{|c|}{\underline{97.81}} &
\multicolumn{1}{c|}{\textbf{97.84}} &
% \multicolumn{1}{c|}{\underline{97.81}} &
\multicolumn{1}{c|}{\underline{97.81}} &
\multicolumn{1}{c|}{97.79} &
% \multicolumn{1}{c|}{96.20} &
\multicolumn{1}{c|}{95.98} \cr 
\cline{1-1}
CIFAR100   &
\multicolumn{1}{|c|}{\textbf{96.41}} &
\multicolumn{1}{c|}{\underline{96.40}} &
% \multicolumn{1}{c|}{96.33} &
\multicolumn{1}{c|}{96.25} &
\multicolumn{1}{c|}{95.18} &
% \multicolumn{1}{c|}{92.42} &
\multicolumn{1}{c|}{91.00} \cr 
\cline{1-1}
FMNIST     &
\multicolumn{1}{|c|}{\textbf{94.19}} &
\multicolumn{1}{c|}{\underline{94.09}} &
% \multicolumn{1}{c|}{94.00} &
\multicolumn{1}{c|}{93.94} &
\multicolumn{1}{c|}{93.69} &
% \multicolumn{1}{c|}{94.06} &
\multicolumn{1}{c|}{93.04} \cr 
\cline{1-1}
CatsVsDogs &
\multicolumn{1}{|c|}{99.59} &
\multicolumn{1}{c|}{\textbf{99.63}} &
% \multicolumn{1}{c|}{\textbf{99.63}} &
\multicolumn{1}{c|}{\underline{99.60}} &
\multicolumn{1}{c|}{\textbf{99.63}} &
% \multicolumn{1}{c|}{98.67} &
\multicolumn{1}{c|}{98.97} \cr
\cline{1-1}
DIOR &
\multicolumn{1}{|c|}{91.74} &
\multicolumn{1}{c|}{\underline{92.52}} &
% \multicolumn{1}{c|}{\underline{93.03}} &
\multicolumn{1}{c|}{\textbf{93.97}} &
\multicolumn{1}{c|}{91.27} &
% \multicolumn{1}{c|}{88.68} &
\multicolumn{1}{c|}{88.78} \cr 
\hline
\end{tabular}
\caption{{\bf Modeling functions in the unimodal setting:} We report AUROC results of our algorithm using different modelling functions on the pre-trained features only. In particular, we try $k$-NN with different values of $k$, the number of nearest neighbors. We also test Gaussian Mixture Model (GMM) with varying number of Gaussians. It can be observed that no modelling function is consistently better than the others. This leads us to use a single Gaussian because if offers an attractive trade-off between high accuracy, low memory footprint and fast computation time.
}
\label{table:reduced_unimodal_knn_gm_ablation}
\end{table}

% \input{latex/tables/knn_gmm_ablations}
% \input{latex/tables/reduced_knn_gm_ablation_unimodal}

% We report results in Table~\ref{table:unimodal}.

We further compared our work against some of the leading methods on additional datasets and report results in Table~\ref{table:additional_dataset}. Our method outperforms other methods on most datasets, often by a large margin (over $18\%$ and $12\%$ on ”Blood Cells” and ”Weather Recognition”, respectively). Our method under-performs only slightly on ”Concrete Crack Classification” and ”Covid19”, in which it comes in second. The different characteristics of these datasets, which belong to very different domains, demonstrate the robustness and flexibility of our method.

% \paragraph{Pre-Trained Vs. Fine-Tuned Features:} 
We further analyze the contribution of pre-trained and fine-tuned features to the final outcome. Results are reported in Table~\ref{table:unimodal_bop_ablation}. We also report in Table~\ref{table:unimodal_bop_ablation} the performance of the algorithm using different numbers of teacher-student  blocks. 
In all cases, the data in each feature space is modeled with a single Gaussian.

As shown in the first four columns, each feature space independently yields good results (left column for pre-trained features, middle three columns for various number of teacher-student blocks used to produce the fine-tuned features).
Combined (the rightmost three columns) we report the best results on most datasets. Using the pre-trained features combined with only the last ViT block for the fine-tuned features yields SOTA results in most cases, suggesting a more compact version of our method. We focus on the pre-trained features combined with 10 blocks teacher-student fine-tuned features, as it achieved the best results on most datasets.

We next considered several modeling functions for the pre-trained features, including $k$NN, a single Gaussian, and a Gaussian Mixture Model (GMM). Results, for the pre-trained features only, are found in Table~\ref{table:reduced_unimodal_knn_gm_ablation}. The main observation we draw from this table is that no particular modelling function is consistently better than others. Therefore, we prefer the use of a single Gaussian, which requires less memory to store and is faster to compute.

A single Gaussian is used in order to model teacher-student fine-tuned features as well, based on empirical distribution of those features (see supplemental for details).
% In the supplenentary material, Figure \ref{fig:bopPreds} shows the fine-tuned features of the last block. One can observe that the feature distribution obeys a Gaussian model qualitatively. Therefore, we chose the scoring model to be the likelihood of a single Gaussian distribution.

\paragraph{Multimodal Setting}
\label{multimodal_setting_results}
\begin{table}[]
\small
\centering
\begin{tabular}{|l|l|l|l|l|l|}
\hline
Dataset    & 
\multicolumn{1}{c|}{\thead{ Deep \\ SVDD}} & 
\multicolumn{1}{c|}{DN2} & 
\multicolumn{1}{c|}{PANDA} & 
\multicolumn{1}{c|}{MSAD} & 
\multicolumn{1}{c|}{Ours} \\
\hline
CIFAR10    &
\multicolumn{1}{c|}{50.67} &
\multicolumn{1}{c|}{71.7} &
\multicolumn{1}{c|}{78.5} &
\multicolumn{1}{c|}{\underline{85.3}} &
\multicolumn{1}{c|}{\textbf{90.38}} \\
\cline{1-1}
CIFAR100   &
\multicolumn{1}{c|}{50.75} &
\multicolumn{1}{c|}{\underline{71.0}} &
\multicolumn{1}{c|}{62.47} &
\multicolumn{1}{c|}{67.65} &
\multicolumn{1}{c|}{\textbf{79.80}} \\
\cline{1-1}
FMNIST     & 
\multicolumn{1}{c|}{70.85} &
\multicolumn{1}{c|}{\underline{77.64}} &
\multicolumn{1}{c|}{\textbf{79.45}} &
\multicolumn{1}{c|}{72.26} &
\multicolumn{1}{c|}{72.53} \\
\cline{1-1}
DIOR     &
\multicolumn{1}{c|}{56.71} &
\multicolumn{1}{c|}{81.10} &
\multicolumn{1}{c|}{\underline{86.92}} &
\multicolumn{1}{c|}{\textbf{90.11}} &
\multicolumn{1}{c|}{66.71} \\
\cline{1-1}
\hline
\end{tabular}
\caption{{\bf AUROC score of the multimodal setting:} We compare our method (rightmost column) against the alternative. $\sharp$ taken from \cite{bergman2020deep}, $\natural$ taken from \cite{reiss2021mean} }
\label{table:multimodal}
\end{table}

\begin{table}[]
\centering
\begin{tabular}{|l|l|l|l|l|l|}
\hline
Dataset    &
\multicolumn{1}{l|}{\thead{$k$-NN \\ $k=2$ }} &
\multicolumn{1}{l|}{\thead{$k$-NN \\ $k=5$ }} &
% \multicolumn{1}{l|}{\thead{$k$-NN \\ $k=10$ }} &
\multicolumn{1}{l|}{\thead{GMM \\ $n=1$ }} &
\multicolumn{1}{l|}{\thead{GMM \\ $n=5$ }} &
% \multicolumn{1}{l|}{\thead{GMM \\ $n=10$ }} &
\multicolumn{1}{l|}{\thead{GMM \\ $n=20$ }} \\
\hline
CIFAR10    &
\multicolumn{1}{|c|}{88.76} &
\multicolumn{1}{c|}{89.16} &
% \multicolumn{1}{c|}{89.40} &
\multicolumn{1}{c|}{90.23} &
\multicolumn{1}{c|}{\textbf{90.81}} &
% \multicolumn{1}{c|}{\underline{90.56}} &
\multicolumn{1}{c|}{\underline{90.39}} \cr 
\cline{1-1}
CIFAR100   &
\multicolumn{1}{|c|}{\underline{82.20}} &
\multicolumn{1}{c|}{\textbf{82.68}} &
% \multicolumn{1}{c|}{\textbf{82.87}} &
\multicolumn{1}{c|}{78.76} &
\multicolumn{1}{c|}{79.42} &
% \multicolumn{1}{c|}{78.99} &
\multicolumn{1}{c|}{77.66} \cr 
\cline{1-1}
FMNIST     &
\multicolumn{1}{|c|}{\underline{75.59}} &
\multicolumn{1}{c|}{74.99} &
% \multicolumn{1}{c|}{74.51} &
\multicolumn{1}{c|}{72.29} &
\multicolumn{1}{c|}{75.43} &
% \multicolumn{1}{c|}{\underline{76.92}} &
\multicolumn{1}{c|}{\textbf{78.00}} \cr 
\cline{1-1}
DIOR &
\multicolumn{1}{|c|}{66.66} &
\multicolumn{1}{c|}{66.08} &
% \multicolumn{1}{c|}{65.73} &
\multicolumn{1}{c|}{65.72} &
\multicolumn{1}{c|}{\underline{69.75}} &
% \multicolumn{1}{c|}{\textbf{70.21}} &
\multicolumn{1}{c|}{\textbf{69.82}} \cr 
\hline
\end{tabular}
\caption{{\bf Modeling functions in the multimodal setting:} We report AUROC results of our algorithm using different modelling functions on the pre-trained features. In particular, we try $k$-NN with different values of $k$, the number of nearest neighbors. We also test Gaussian Mixture Model (GMM) with varying number of Gaussians. It can be observed that no modelling function is consistently better than the others. This leads us to use a single Gaussian because if offers an attractive trade-off between high accuracy, low memory footprint and fast compute time. 
% \matan{1. This is a copy of the description of Table 4. Maybe we should merge them? 2. Shir's comment:"GMM, N=1 is not the best in any category and even the worst on two of the four datasets. Im semsomg this could be problematic.". She's got a point...}
}
\label{table:reduced_multimodal_knn_gm_ablation}
\end{table}

\begin{table}
\centering
\begin{tabular}{|c|c|c|c|c|}
\hline
 \thead{Explained \\ Variance}    &
\multicolumn{2}{c|}{CIFAR10} &
\multicolumn{2}{c|}{Weather Recognition} \\
\hline
\multicolumn{1}{|l|}{\centering {}} &\multicolumn{1}{l|}{\centering {Pre-Trained}} & \multicolumn{1}{l|}{\centering {Full}} &\multicolumn{1}{l|}{\centering {Pre-Trained}} & \multicolumn{1}{l|}{\centering {Full}} \\
\hline
 $85\%$ & 96.85 & 98.11 & 81.46 & 94.43  \\
\hline 
 $90\%$ & 97.81 & 98.31 & 81.06 & 94.32 \\
\hline
 $95\%$ & 98.11 & 98.33 & 81.45 & 94.21\\
\hline
\end{tabular}
\caption{{\bf Sensitivity of Whitening hyperparameter:} We whiten and reduce the dimensionality of the pre-trained features by keeping $90\%$ of the energy. In this experiment, we show the result of using either $85\%$ or $95\%$ of the energy on two different datasets. The left column, in each table, shows the performance of only the pre-trained features. The right column shows the performance of the full algorithm (pre-trained + fine-tuned features). As can be seen, the algorithm is not sensitive to this hyperparameter.}
\label{table:combined_whiten_var}
\end{table}
We further tested our algorithm in the multimodal setting, where one of the classes is considered abnormal while all other classes are considered normal. That is, all samples of the normal classes are used as single multimodal class, without using their original labels. 
% With AUROC as our evaluation metric, we compare our model with the leading pre-trained models.

We report the AUROC results in Table~\ref{table:multimodal}. As can be seen, the proposed method achieved SOTA results on cifar10 (AUROC score of $90.24$) and Cifar100 (AUROC score of $83.05$), outperforming alternative methods by approximately $5\%$ and $12\%$, respectively. 

The performance of our method degrades when using the grayscale Fashion MNIST dataset. We suspect that this might be due to the fact that the grayscale dataset is not aligned with the pretraining phase of ViT, which used color images. 

We observe a sharp drop in the performance of our method on the DIOR dataset when switching from the unimodal to the multimodal setting. We thoroughly discuss the details of this drop in sub-section ~\ref{limitations.}.  

Interestingly, in the multimodal setting, the performance of the algorithm does not change much as we try different modeling functions, see Table~\ref{table:reduced_multimodal_knn_gm_ablation}. We observe that using $k$NN (with different values of $k$), as well as a Gaussian Mixture Model (GMM) with varying number of Gaussians gives similar results.

\paragraph{Whitening:} Finally, in the last experiment we test the robustness of our algorithm to the dimensionality reduction parameter. Since we use a Gaussian with full covariance to model the pre-trained features, we reduce the dimensionality of the data and improve its structure by whitening it first and keeping enough dimensions to preserve $90\%$ of energy.

We have tried other thresholds ($85\%$ and $95\%$) and, as shown in Table~\ref{table:combined_whiten_var}, our method performed well with all thresholds, demonstrating that our method is not sensitive to this hyperparameter's choice. One can observe that the fine-tuned features boost performance using all thresholds, and on "Weather Recognition" by a large margin.

\subsection{Limitations}
\label{limitations.}

% \begin{figure}
% \begin{center}
% \includegraphics[width=1\linewidth]{latex/figures/unimodal_vs_multimodal.png}

% \end{center}
%   \caption{"Pre-training Confusion" in synthetic Setting. Each shape represents a different class in the dataset. In a Unimodal setting only squares are considered normal. In a multimodal setting all classes except squares are considered normal. The situation is not symmetric and affects AUROC scores. See text for details.}
% \label{fig:unimodal_vs_multimodal}
% \end{figure}

% \begin{figure*}[h!]
%     \centering
%     \begin{tabular}{cc}
%     \includegraphics[width=0.5\textwidth]{latex/figures/class_13_pretrained_ViT_testset.png} &
%     \includegraphics[width=0.5\textwidth]{latex/figures/cifar10_resnet_embeddings.png} \\
%     \end{tabular}
%     \caption{(Left) tSNE of pre-trained ViT penultimate layer outputs of DIOR. As can be seen, Although the samples embedding are semantically separated for some classes, samples embeddings of classes 13 and 17 are mixed as well as samples embeddings of class 4 and class 8. (Right) tSNE of pre-trained ResNet penultimate layer outputs of Cifar10. As can be seen, Although the samples embedding are semantically separated for some classes, samples embeddings of classes 3 and 5 are mixed.
%     }
%     \label{fig:confusion}
% \end{figure*}

\begin{figure*}[ht!]
   \subfloat[Synthetic Setting \label{fig:toy_example}]{%
      \includegraphics[ width=0.31\textwidth]{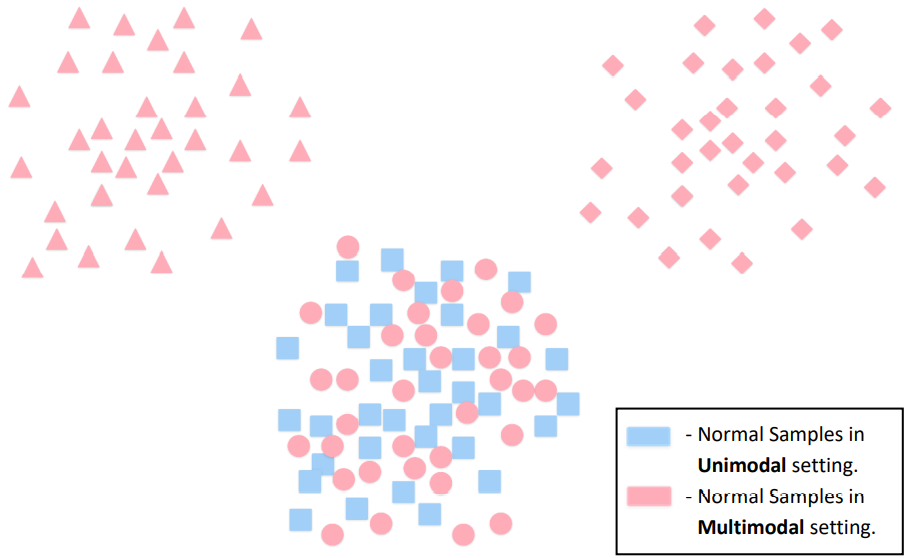}}
\hspace{\fill}
   \subfloat[ViT DIOR Embeddings \label{fig:vit_confusion} ]{%
      \includegraphics[ width=0.34\textwidth]{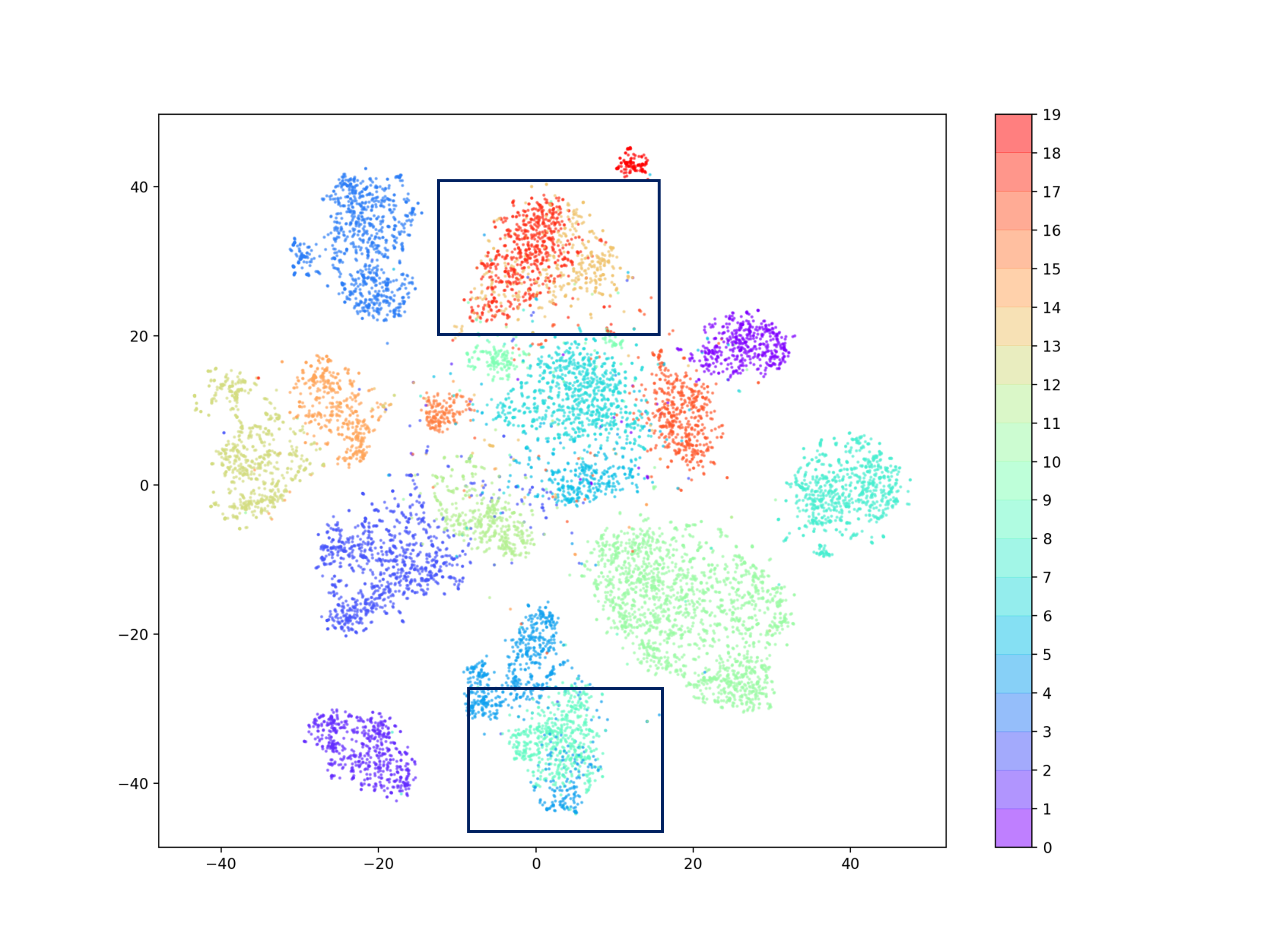}}
\hspace{\fill}
   \subfloat[ResNet CIFAR10 Embeddings \label{fig:resnet_confusion}]{%
      \includegraphics[ width=0.33\textwidth]{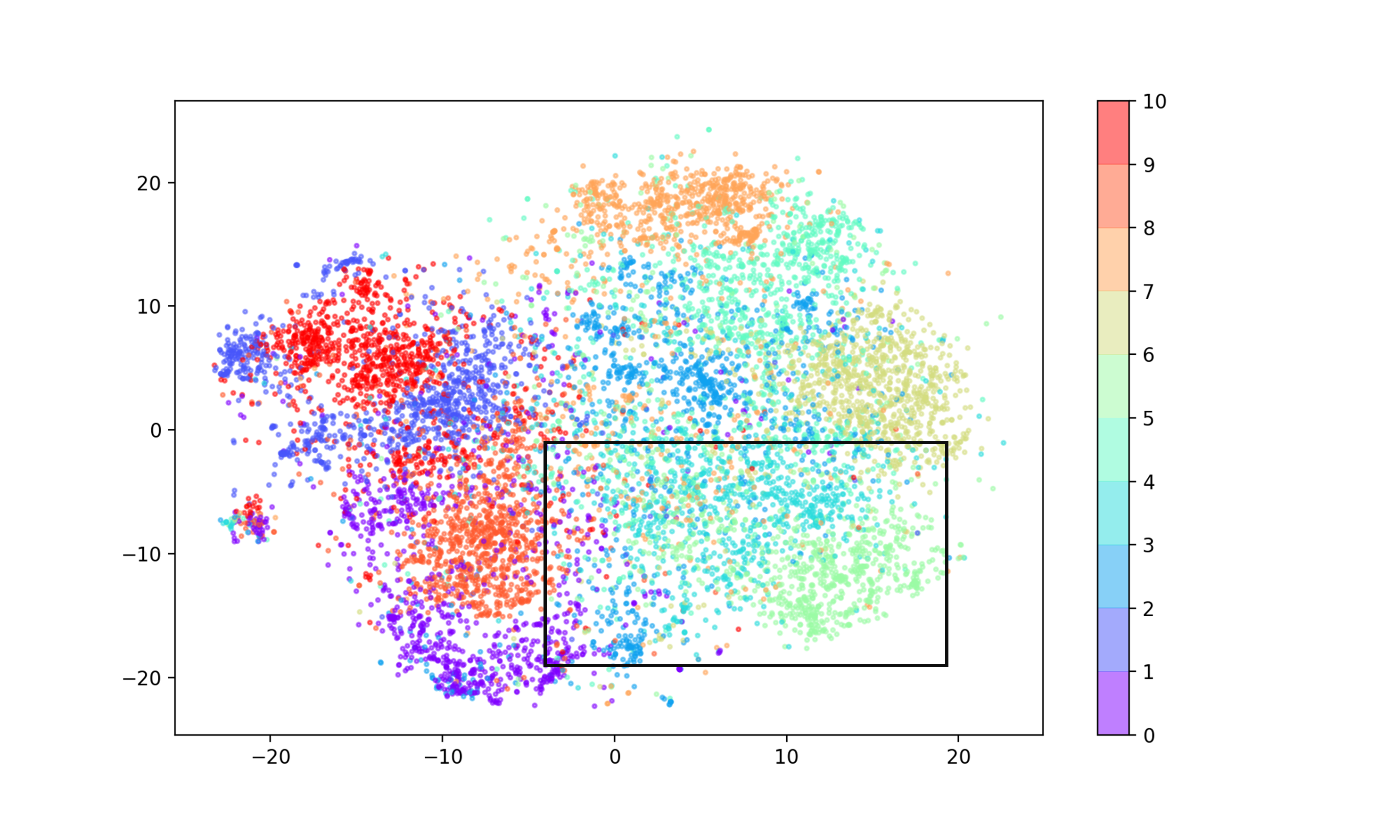}}\\
\caption{{\bf"Pre-training Confusion" in synthetic Setting, DIOR ViT embeddings and CIFAR10 ResNet embeddings:} (a) Synthetic Setting - each shape represents a different class in the dataset. In a unimodal setting only squares are considered normal. In a multimodal setting all classes except squares are considered normal. The situation is not symmetric and affects AUROC scores. See text for details. (b) tSNE of pre-trained ViT penultimate layer outputs of DIOR. As can be seen, although the sample embeddings are semantically separated for some classes, sample embeddings of classes 13 and 17 are mixed as well as sample embeddings of class 4 and class 8. (c) tSNE of pre-trained ResNet penultimate layer outputs of Cifar10. As can be seen, although the sample embeddings are semantically separated for some classes, sample embeddings of classes 3 and 5 are mixed. Best viewed in color. Zoom in for details. }
    \label{pretriaining_confusion}
\end{figure*}

Evaluating Transformaly in both the unimodal and multimodal settings reveals the strengths and limitations of our method and the pre-training approach. We outperform almost all methods in the unimodal case, and achieve SOTA results on Cifar10 and Cifar100, in the multimodal case. However, in the multimodal case we do fail on the DIOR dataset. This occurs because of "pre-training confusion", where the pre-trained model maps two semantically different classes to the same region in feature space.

Figure~\ref{fig:toy_example} shows a toy example of this effect in the case of four semantically different classes (triangles, diamonds, circles, and squares). The triangles and diamonds are nicely separated, while the squares and circles are confused. 

Consider the unimodal case, where only the blue squares are available as normal samples during training. In this case, at test time only the red circles will be confused as normal instead of abnormal. The red triangles and diamonds will be correctly classified as abnormal. The algorithm misses {\em some} of the abnormalities.

The situation is reversed in the multimodal case. Assume now that all red samples (triangles, diamonds, and circles) are normal. At test time, all the abnormal blue squares will be classified as normal. The algorithm misses {\em all} the abnormalities.

We suspect that the presented pre-training confusion happens in the DIOR case. 
To validate this, we plot a tSNE embedding of the features of DIOR in  Figure~\ref{fig:vit_confusion}. One can observe that our pre-trained model confuses between class 13 and class 17 and between class 4 and class 8 (highlighted). That is, the model produces embeddings that are similar for both classes.

A similar phenomenon occurs with a ResNet architecture as well. This might explain the failure of recently suggested ResNet-based methods on Cifar10 and Cifar100 in the multimodal setting (such as DN2 \cite{bergman2020deep} and PANDA\cite{reiss2020panda}). A tSNE embedding of the pre-trained ResNet features of Cifar10 is plotted in Figure~\ref{fig:resnet_confusion}. One can observe that pre-trained ResNet model confuses between class 3 and class 5 (highlighted). 

The stress testing of anomaly detection algorithms in the multimodal settings helps to reveal their limitations. We believe that further analyzing anomaly detection in the multimodal setting is an important topic for future research.

% ##################################

% In this scenario, one might wonder if the teacher-student features will assist in anomaly detection. Unfortunately, the fact that our teacher architecture is the pre-trained architecture, which mixes the outputs of the two classes, will result in similar behavior from the student blocks. Therefore the teacher-student features will not be suitable for anomaly detection.
%  \matan{I think the teacher-student mentioned here will lead to more questions from the reviewers. I think I would remove the last two sentences and answer it if the reviewers will ask in the rebuttal.}

% One of our motivations for the multimodal evaluation is this presented phenomenon. One might argue that adding a precision score to standard evaluation process will detect this silent confusion failure. Although it can reveal this delicate situation, it will force the future suggested methods to set a threshold in order to hard-classify samples, which is an undesirable evaluation characteristic. Furthermore, This suggestion does not address the previously mentioned Unimodal limitations, which can be found in in the begining of this section. Thus, we believe anomaly detectors should be evaluated using both the Unimodal and the multimodal settings. Furthermore, fully addressing the "pre-training confusion" failure in the multimodal setting could be a topic for future research.

\section{Conclusions}

Transformaly is an anomaly detection algorithm that is based on the Visual Transformer (ViT) architecture. The data is mapped to a pre-trained feature space and a fine-tuned feature space. The normality score of a query point is based on the product of its likelihood in both spaces. 
% Previous work used either pre-trained features, or fine-tuned features, but not both.

Pre-trained features are obtained by running the samples through a pre-trained ViT. Fine-tuned features are obtained by training a student-network, on normal data only. The discrepancy between student and teacher networks forms the fine-tuned features.

We conduct extensive experiments on multiple datasets and obtain consistently good results, often surpassing the current state of the art.
{\small
\bibliographystyle{ieee_fullname}
\bibliography{latex/ms}
}

\section{Appendix}
In this section we will further explain the datasets we used, present the gaussian nuture of the pre-trained features and demonstrate our method's robustness.

  \subsection{Datasets Details}
  \label{datasets_details}
 
\textbf{CIFAR} consists of two well known datasets, Cifar10 and Cifar100, that are used for various tasks including semantic anomaly detection \cite{krizhevsky2009learning}. Each dataset contains $60,000$ $32 \times 32$ color natural images, split into $50,000$ training images and $10,000$ test images. Cifar10 is composed of 10 equal-sized classes, whereas cifar100 has 100 equal-sized fine-grained classes or 20 equal-sized coarse-grained classes. Following the previous papers, we use the coarse-grained classes notation.

\textbf{Fashion MNIST} consists of $60,000$ train samples and $10,000$ examples test samples ~\cite{xiao2017fashion}. Each example is a $28 \times 28$ grayscale image labeled with one of 10 different categories. 

\textbf{Cats Vs Dogs} is a dataset of images of cats and dogs. The training set contains $10,000$ images of cats and $10,000$ images of dogs, while the test set contains $2,500$ dog images and $2,500$ cat images. There is either a dog or a cat in every image, appearing in a variety of poses and scenes. Following previous work~\cite{bergman2020deep,reiss2020panda}, we split each class to the first $10,000$ images for training and the last 2,500 for testing.

\textbf{Dior} contains aerial images with 19 object categories. Following previous papers ~\cite{bergman2020deep,reiss2020panda}, we used the bounding boxes provided with the data, and we took objects with at least 120 pixels in each axis as well as only classes with more than 50 images. This preprocessing phase led to 19 classes, with an average training size of 649 images. The sample sizes in each class are not equal, as the lowest sample size in the training set is 116 and the highest is 1890.

\textbf{Blood Cells}~\cite{shenggan_bccd_2021} contains $320 \times 240$ augmented color images of four different cell types. The training set contains approximately $2,500$ images for each blood cell type, whereas the test set contains approximatly $620$ images for each type of blood cell.

\textbf{Covid19}~\cite{noauthor_chest_nodate} is a dataset of Chest X-ray images of Covid19, Pneumonia and normal patients.We ignore the Pneumonia patients' scans and have used just the Covid19 and normal scans. Covid19 patients' chest X-rays have been divided into $460$ images in the training set and $116$ images in the test set. The chest X-ray images of normal patients have been divided into $1,266$ images for the training set and $317$ images for the test set. Normal patients' scans are obviously considered normal, while Covid19 patients' scans are considered anomalous.

\textbf{View Recognition}~\cite{noauthor_intel_nodate} is an image dataset of natural scenes around the world. This dataset is composed of six different classes such as images of forest and streets. The training set contains approximately $2,300$ images for each class, while the test set contains approximatly $500$ images for each class.

\textbf{Weather Recognition}~\cite{ajayi_multi-class_2018} is a multi-class dataset of weather images designed for image classification. There are four types of outdoor weather images in this dataset, including shine and rain. The training set consists of approximately $225$ images per class, while the test set contains approximately $55$ images per class.

\textbf{Concrete Crack Classification}~\cite{ozgenel_concrete_2019} contains $227 \times 227$ color concrete images with and without cracks. There are $16,000$ images per class in the training set and $4,000$ per class images in the test set. Images of concrete without cracks are considered normal, while images of concrete with cracks are considered anomalous.

\subsection{Gaussian nature of data}
\label{gaussian_nature}
\begin{figure}[h]
\begin{center}
\includegraphics[width=1\linewidth]{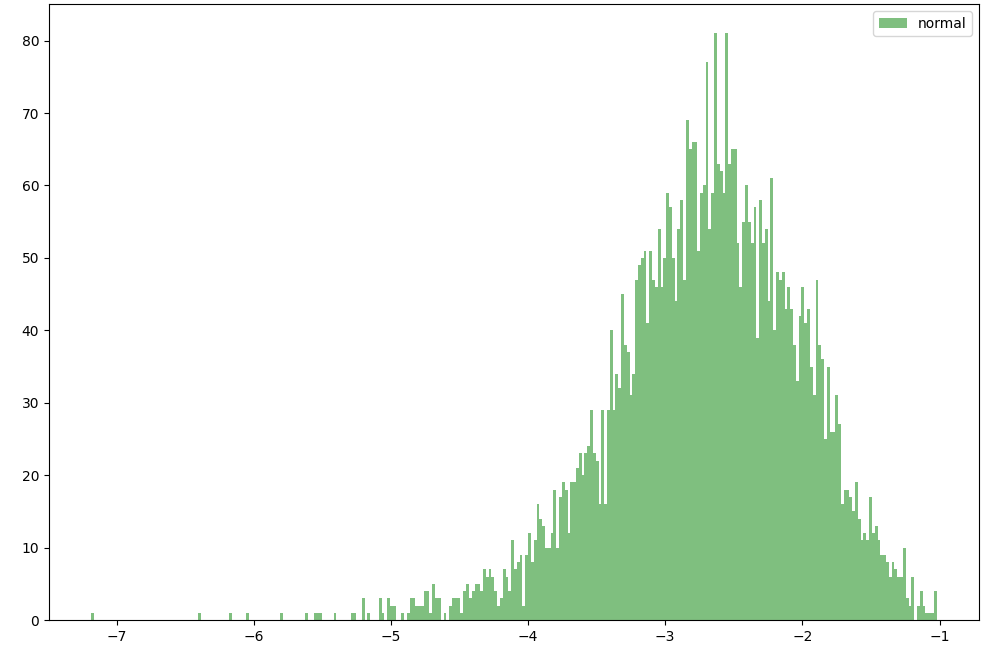}
\end{center}
  \caption{{\bf Gaussian nature of data:} We show here the Teacher-Student fine-tuned features from the last ViT block using class 0 samples as the normal training set. This distribution can easily be fitted with a Gaussian model, explaining the good results we get using this module. The behaviour of other fine-tuned features of other classes is similar.}
\label{fig:bopPreds}
\end{figure}

In this section, we presents the fine-tuned feature empirical distribution, that explains why a Gaussian is used to model this data. Figure~\ref{fig:bopPreds} shows the Teacher-Student fine-tuned features of the last ViT block, using class 0 samples as the normal training set. As one can observe, the fine-tuned features follow a distribution close to Gaussian, which motivate us to use a Gaussian to model the data. We observed similar empirical distributions using different ViT blocks and other normal classes.
 
 \subsection{Transformaly Robust Results}
 To further demonstrate our robust results, we repeated our unimodal experiments three times. In each of the three trials, we calculate the average AUROC score across all possible class choices. Table \ref{table:unimodal_with_sd} shows the mean and standard deviation scores of our method, calculated over these three trails. Transformaly achieved similar results, still outperforming other methods on all datasets, except for FMNIST.

\begin{table}[]
\small
\centering
\begin{tabular}{|l|l|l|l|l|l|}
\hline
Dataset    &
\multicolumn{1}{l|}{CSI} &
\multicolumn{1}{l|}{DN2} &
\multicolumn{1}{l|}{PANDA} &
\multicolumn{1}{l|}{MSAD} &
\multicolumn{1}{l|}{Ours} \\
\hline
CIFAR10    &
\multicolumn{1}{c|}{94.3} &
\multicolumn{1}{c|}{92.5} &
\multicolumn{1}{c|}{96.2} &
\multicolumn{1}{c|}{97.2} &
\multicolumn{1}{c|}{\textbf{98.34($\pm$ 0.018)}} \cr 
% \cline{1-1}
\hline
CIFAR100   &
\multicolumn{1}{c|}{89.6} &
\multicolumn{1}{c|}{94.1} &
\multicolumn{1}{c|}{94.1} &
\multicolumn{1}{c|}{96.4} &
\multicolumn{1}{c|}{\textbf{97.60($\pm$ 0.184)}} \cr
\hline
FMNIST    &
\multicolumn{1}{c|}{-} &
\multicolumn{1}{c|}{94.5} &
\multicolumn{1}{c|}{\textbf{95.6}} &
\multicolumn{1}{c|}{94.21} &
\multicolumn{1}{c|}{94.37($\pm$ 0.041)} \cr 
\cline{1-1}
\hline
CatsVsDogs    &
\multicolumn{1}{c|}{86.3} &
\multicolumn{1}{c|}{96.0} &
\multicolumn{1}{c|}{97.3} &
\multicolumn{1}{c|}{99.3} &
\multicolumn{1}{c|}{\textbf{99.47($\pm$ 0.037)}} \cr
% \cline{1-1}
\hline
DIOR    &
\multicolumn{1}{c|}{78.5} &
\multicolumn{1}{c|}{92.2} &
\multicolumn{1}{c|}{94.3} &
\multicolumn{1}{c|}{97.2} &
\multicolumn{1}{c|}{\textbf{98.33($\pm$ 0.177)}} \cr
\hline
\end{tabular}
\caption{{\bf AUROC scores of the unimodal setting:} In each trial we calculate the mean AUROC score across all classes of the datasets. We repeat this process for three trials reporting its means and standard deviations. Other benchmarks's AUROC scores are copied from the original table.}
\label{table:unimodal_with_sd}
\end{table}

\end{document}